\documentclass[runningheads]{llncs}

\usepackage{eccv}

\usepackage{eccvabbrv}

\usepackage{graphicx}
\usepackage{booktabs}
\usepackage{wrapfig}

\usepackage[accsupp]{axessibility}  %

\usepackage[hidelinks,colorlinks=true]{hyperref}

\usepackage{orcidlink}

\usepackage{tcolorbox}
\usepackage{fontawesome5}
\usepackage{xcolor}

\newtcbox{\orangebutton}{on line,
  colback=orange,
  colframe=orange,
  coltext=white,
  arc=3pt,
  boxrule=0pt,
  left=5pt,right=5pt,top=2.5pt,bottom=2.5pt}

\newtcbox{\blackbutton}{on line,
  colback=black,
  colframe=black,
  coltext=white,
  arc=3pt,
  boxrule=0pt,
  left=5pt,right=5pt,top=2.5pt,bottom=2.5pt}

\begin{document}

\title{UniMesh: Unifying 3D Mesh Understanding and Generation} 

\titlerunning{UniMesh: Unifying 3D Mesh Understanding and Generation}

\author{
  \textbf{Peng Huang$^{1*}$~~
  Yifeng Chen$^{2*}$~~
  Zeyu Zhang$^{2*\dag}$~~
  Hao Tang$^{2\ddag}$}\\
  \vspace{.5em}
  $^{1}$Boston University~~
  $^{2}$Peking University\\
  \vspace{.5em}
  \small$^{*}$Equal contribution. $^{\dag}$Project lead. $^{\ddag}$Corresponding author: bjdxtanghao@gmail.com.
}

\authorrunning{Peng Huang and Yifeng Chen et al.}

\institute{}

\maketitle

\begin{center}
\href{https://aigeeksgroup.github.io/UniMesh}{
  \orangebutton{\faGlobe\ \textbf{Website}}
}
\hspace{8pt}
\href{https://github.com/AIGeeksGroup/UniMesh}{
  \blackbutton{\faGithub\ ~~~\textbf{Code}}
}
\end{center}

\begin{figure*}[h] %
	\centering
    \includegraphics[width=\textwidth]{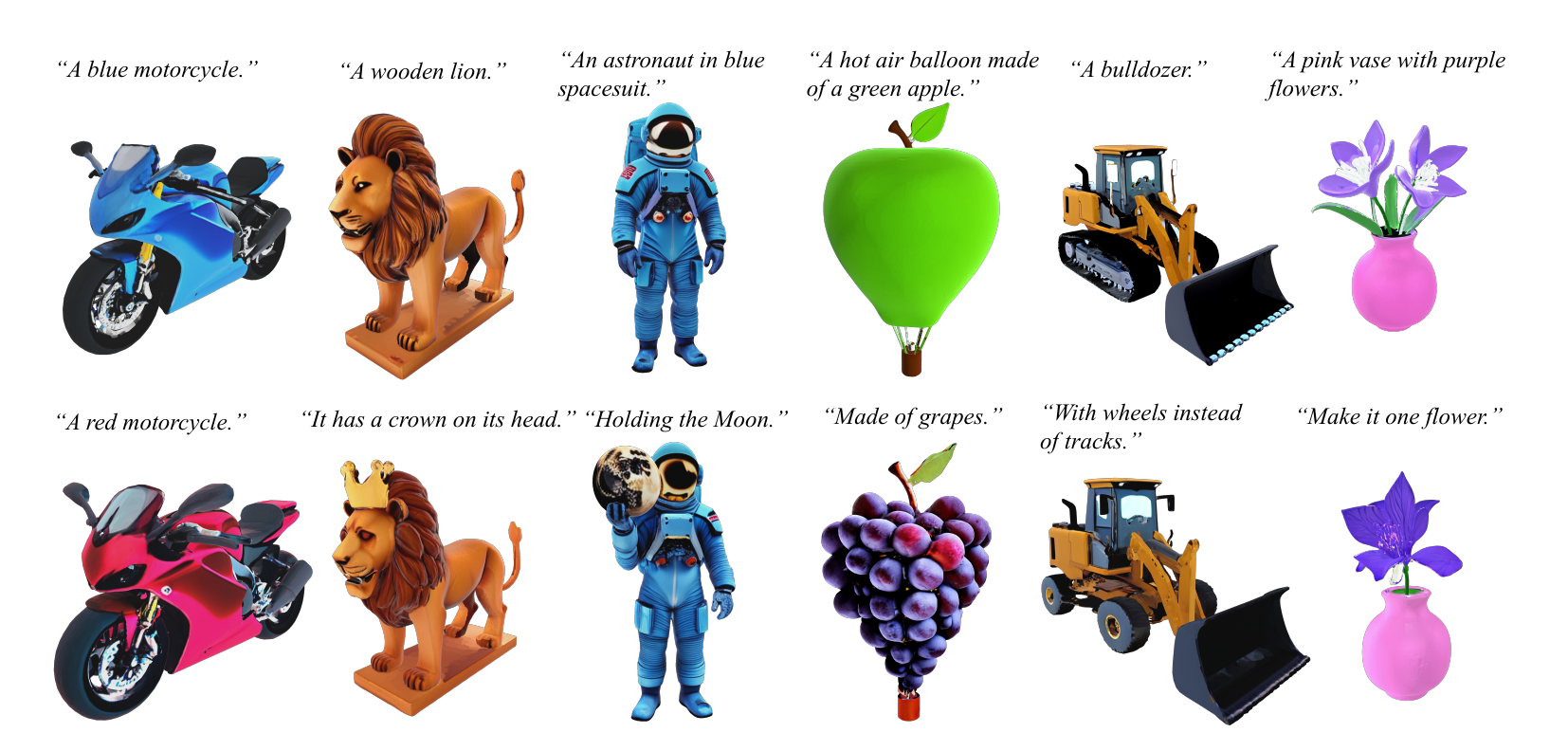}
    \captionof{figure}{ \textbf{UniMesh enables semantic-aware 3D mesh generation and editing.} From a single text prompt (top row), UniMesh generates high-fidelity 3D meshes. Leveraging its unified understanding--generation architecture, it further supports iterative semantic edits (bottom row), such as changing object color (``blue motorcycle'' $\rightarrow$ ``red motorcycle''), adding attributes (``astronaut'' $\rightarrow$ ``astronaut holding the Moon''), or modifying structure (``tracks'' $\rightarrow$ ``wheels''), demonstrating the synergy between 3D understanding and generation capabilities within the \textit{Chain-of-Mesh} mechanism.}
    \label{fig:teaser}
\end{figure*}

\begin{abstract}
  Recent advances in 3D vision have led to specialized models for either 3D understanding (e.g., shape classification, segmentation, reconstruction) or 3D generation (e.g., synthesis, completion, and editing). However, these tasks are often tackled in isolation, resulting in fragmented architectures and representations that hinder knowledge transfer and holistic scene modeling. To address these challenges, we propose \textbf{UniMesh}, a unified framework that jointly learns 3D generation and understanding within a single architecture. First, we introduce a novel \textbf{Mesh Head} that acts as a cross-model interface, bridging diffusion-based image generation with implicit shape decoders. Second, we develop \textbf{Chain-of-Mesh (CoM)}, a geometric instantiation of iterative reasoning that enables user-driven semantic mesh editing through a closed-loop "latent, prompting, and re-generation" cycle. Third, we incorporate a \textbf{self-reflection} mechanism based on an Actor–Evaluator–Self-reflection triad to diagnose and correct failures in high-level tasks like 3D captioning. Experimental results demonstrate that UniMesh not only achieves competitive performance on standard benchmarks but also unlocks novel capabilities in iterative editing and mutual enhancement between generation and understanding.

\end{abstract}

\section{Introduction}
\label{sec:intro}

Recent advances in 3D vision have led to a flourishing of specialized models. On one side, large reconstruction models (LRMs) \cite{hong2024lrmlargereconstructionmodel} like InstantMesh \cite{xu2024instantmeshefficient3dmesh} and Hunyuan3D \cite{hunyuan3d2025hunyuan3d} can synthesize high-fidelity 3D assets from minimal inputs in seconds~\cite{zhang2026code2worlds}. On the other side, sophisticated understanding systems have reached high levels of proficiency in semantic reasoning, part segmentation \cite{nguyen20133dseg, woo2002new, xu2020squeezesegv3}, and 3D captioning~\cite{zhao2026cov,huang2025dc,tang20263d,huang20253d,liu2025nav}.

Despite these individual successes, 3D generation and understanding remain fundamentally fragmented~\cite{duan2026liveworld,wu2026light4d,wang2026safemo,zhang2024infinimotion,zhang2024motion,zhang2024kmm,zhang2025flashmo,ouyang2025motion}. They typically reside in isolated architectures with incompatible representations. This dichotomy prevents meaningful knowledge transfer between the two fields and hinders the development of holistic scene modeling, where a system should be able to both create content and reason about its own outputs to improve them.

To address these challenges, we need to create a unified framework that couples perception and creation. By moving away from "one-pass" generation models—which cannot inherently support iterative, user-driven semantic edits—we aim to build a system capable of a closed-loop "latent, prompting, and re-generation" cycle. This mimics the iterative reasoning found in language models, applying it to geometric and semantic 3D tasks.

Hence in this work, we propose \textbf{UniMesh}, the first unified framework that jointly learns and iteratively couples 3D generation and understanding within a single, cohesive architecture. At the core of UniMesh is a novel \textbf{Mesh Head} — a cross-model interface that seamlessly bridges BAGEL’s diffusion-based image generation pipeline and Hunyuan3D’s implicit shape decoder. Then we introduce \textbf{Chain-of-Mesh (CoM)}, a geometric instantiation of iterative reasoning inspired by language-based Chain-of-Thought \cite{wei2022chain}. In CoM, the reference image latent of the generated 3D mesh — along with a new editing prompt — is fed into BAGEL’s Qwen module. By jointly processing the visual context of the current mesh and the textual instruction, Qwen generates an updated image latent that captures the desired modification. This latent is then translated by the Mesh Head into a refined conditioning signal for Hunyuan3D, yielding an edited mesh that aligns with the new prompt. Finally, UniMesh further incorporates a \textbf{self-reflection} \cite{renze2024self} mechanism inspired by recent language-agent frameworks. For high-level tasks like 3D captioning, the system simulates an Actor–Evaluator–Self-reflection triad: if the initial caption is deemed incorrect, a verbal reflection is synthesized to diagnose the failure and guide regeneration, significantly improving the model’s performance on 3D understanding tasks. We conduct extensive experiments, which demonstrate that UniMesh achieves competitive performance across both 3D generation and understanding benchmarks. More importantly, it unlocks capabilities that are absent in conventional models: iterative mesh editing via CoM, mutual enhancement between generation and understanding, and few-shot error correction via self-reflection. We believe UniMesh represents a significant step toward holistic 3D intelligence—where models not only generate but also comprehend, critique, and improve their own 3D creations.

In summary, our contributions can be listed in three folds: 
\begin{enumerate}
    \item We unify BAGEL and Hunyuan3D into a single pipeline via a novel  \textbf{Mesh Head} module — which directly maps image latents from BAGEL’s Qwen to conditioning latents compatible with Hunyuan3D’s shape generator — thereby preserving geometric fidelity while bypassing lossy RGB reconstruction. 
    \item We propose \textbf{Chain-of-Mesh (CoM)} , an iterative refinement strategy that leverages the understanding capability to guide the generative process toward semantically coherent edits — enabling zero-shot, text-guided 3D object modification without retraining. We also introduce a \textbf{self-reflection} mechanism inspired by recent findings in large language models, where the model iteratively evaluates its own predictions, identifies inconsistencies in 3D understanding, and refines its internal representations to improve downstream performance. 
    \item We conduct extensive experiments, which demonstrate that \textbf{UniMesh} achieves promising results across multiple 3D understanding and generation benchmarks, while enabling mutual enhancement between the two modalities — showcasing the benefits of a truly unified 3D vision system.
\end{enumerate}

\begin{figure}[htbp]
	\centering
	\includegraphics[width=\textwidth]{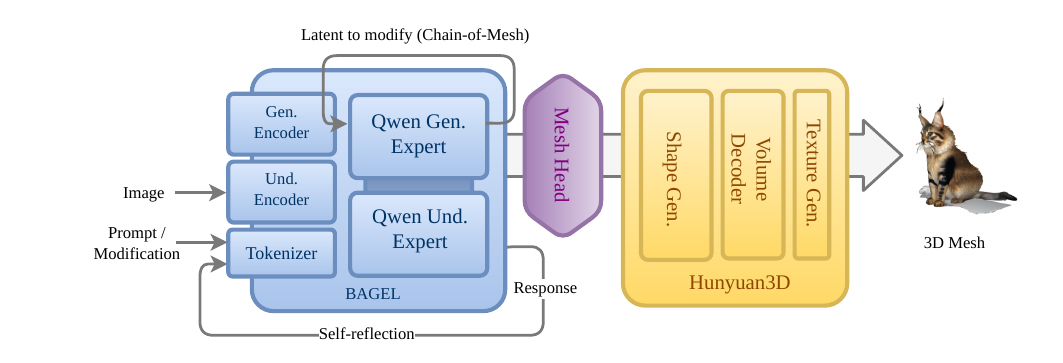}
	\caption{\textbf{Framework of UniMesh.} Given a text prompt or modification instruction, BAGEL with Qwen generates an image latent, which is transformed by the Mesh Head into a conditioning latent for Hunyuan3D to produce a 3D mesh. The reference image latent of the generated mesh can be fed back into BAGEL for iterative refinement via Chain-of-Mesh, while self-reflection enables semantic feedback loops for understanding tasks.}
	\label{fig:framework}
\end{figure}

\section{Related Work}

\subsection{3D Object Generation}

Recent years have witnessed remarkable progress in feed-forward 3D generation~\cite{shi2025revisiting,wang2025zpressor,wang2025volsplat,wang2025drivegen3d,zhang2025vasevqa,zhang2025dragmesh,li2026partrag,zhang2024motionavatar,zhang2025motion}, shifting from slow optimization-based methods to real-time, large-scale generative systems. Early efforts focused on category-specific reconstruction from single images, but the advent of large vision-language models and multi-view diffusion priors has enabled generalizable, high-fidelity 3D synthesis from diverse inputs. 

A key line of work leverages large reconstruction models trained on massive 3D datasets. LRM pioneers this direction by employing a 500M-parameter transformer \cite{vaswani2017attention} to directly regress a NeRF \cite{mildenhall2021nerf} from a single image, trained end-to-end on about 1M objects from Objaverse \cite{objaverse} and MVImgNet \cite{yu2023mvimgnet}. Building upon this architecture, TripoSR \cite{tochilkin2024triposr} further refines data processing and training strategies to achieve sub-second mesh generation with improved geometric fidelity. Similarly, Hunyuan3D-2.1 \cite{hunyuan3d2025hunyuan3d} introduces a modular pipeline comprising a DiT-based \cite{peebles2023scalable_dit} shape generator and a texture synthesizer, enabling high-resolution textured mesh creation tailored for industrial applications. 

Parallel efforts explore alternative 3D representations beyond NeRFs. Large Multi-View Gaussian Model (LGM) \cite{tang2024lgm} proposes multi-view Gaussian features as an efficient, differentiable representation, coupled with an asymmetric U-Net \cite{ronneberger2015unet} backbone to generate 512-resolution 3D Gaussians \cite{kerbl20233dgs} in seconds. Large Gaussian Reconstruction Model (GRM) \cite{xu2024grm}  extends this idea to sparse-view reconstruction, predicting pixel-aligned 3D Gaussians via a transformer in just 0.1s, demonstrating strong transferability to generative tasks when combined with multi-view diffusion models. 

To bridge diffusion priors and explicit geometry, several methods integrate off-the-shelf 2D/3D models into unified pipelines. InstantMesh \cite{xu2024instantmeshefficient3dmesh} synergizes a multi-view diffusion model with an LRM-style reconstruction head, incorporating differentiable iso-surface extraction to directly supervise mesh outputs—enabling scalable training with geometric cues like depth and normals. SF3D \cite{sf3d2024} takes a mesh-centric approach, explicitly predicting UV-mapped textures, material parameters, and normal maps, along with a delighting module to ensure relighting compatibility. Meanwhile, SV3D \cite{voleti2024sv3d} leverages image-to-video diffusion models to generate temporally consistent orbital multi-view videos, which are then lifted to 3D via improved optimization techniques—highlighting the value of multi-view consistency for reconstruction quality. 

While these methods achieve impressive results in either generation or reconstruction, they typically treat 3D understanding (e.g., semantic reasoning, structural analysis) as a separate downstream task. In contrast, UniMesh is a single unified model that jointly supports both 3D content creation—via Chain-of-Mesh for iterative semantic editing—and 3D understanding—via self-reflection for tasks like caption refinement—within one coherent architecture.

\subsection{3D Object Understanding}

Understanding 3D objects through multimodal learning has become a central direction in vision–language research. Early efforts focused on transferring 2D vision–language alignment to 3D by aggregating multi-view visual cues. Recent studies further highlight that 3D understanding requires not only visual appearance comprehension but also shape reasoning, part-level awareness, and the ability to link geometry with natural language semantics.

Cap3D \cite{luo2023scalable} establishes a large-scale 3D captioning pipeline that uses pretrained Vision-Language Models (VLMs) to generate captions for objects from datasets such as Objaverse and ShapeNet. DiffuRank \cite{luo2024view} further extends this idea by introducing diffusion-based view ranking for selecting the most informative views before captioning, thereby improving both descriptive coverage and alignment with ground-truth semantics. LLaVA-3D \cite{zhu2024llava} also achieved advances in 3D model description.
Beyond captioning, recent methods aim to equip models with deeper reasoning about 3D geometry and object structure. 3D-R1 \cite{huang20253dr1} has made a significant contribution in this respect.

Recently, general-purpose VLMs have rapidly expanded their capabilities and increasingly serve as backbones or evaluators for 3D understanding. Open-source models such as DeepSeek-VL2 \cite{wu2024deepseekvl2mixtureofexpertsvisionlanguagemodels}, Molmo-72B \cite{molmo2024}, Qwen2.5-VL \cite{Qwen2.5-VL}, InternVL3.5 \cite{wang2025internvl3_5}, Kimi-VL \cite{kimiteam2025kimivltechnicalreport}, and Phi-4-multimodel \cite{microsoft2025phi4minitechnicalreportcompact} provide strong visual–textual alignment and benefit from large-scale multimodal pretraining, making them adaptable to tasks like multi-view reasoning, shape classification, and 3D captioning. Meanwhile, closed-source VLMs such as Gemini 2.5 Pro \cite{comanici2025gemini25pushingfrontier}, Claude Sonnet 4 \cite{anthropic2025claude4}, and GPT-5 \cite{openai2025gpt5} are also valuable for 3D understanding tasks due to their strong cross-view consistency and world knowledge.

\section{The Proposed Method}

\subsection{Overview}

UniMesh is a unified 3D vision framework that jointly optimizes 3D generation and 3D understanding within a single, cohesive architecture. Unlike conventional pipelines that treat these tasks in isolation—using separate models for shape synthesis and semantic reasoning—UniMesh establishes bidirectional communication between a generative backbone and an understanding module, enabling them to mutually reinforce each other. 

As shown in \cref{fig:framework}, at the heart of UniMesh lies the Mesh Head, a novel cross-model interface that bridges BAGEL’s diffusion-based image generation pipeline and Hunyuan3D’s implicit shape decoder. This module allows direct translation from BAGEL’s image latent space to Hunyuan3D’s conditioning space, bypassing lossy RGB reconstruction and preserving geometric fidelity. The Mesh Head is first trained via supervised fine-tuning on large-scale 3D data, ensuring accurate one-step shape prediction from single-view inputs. 

Beyond static generation, UniMesh enables semantic 3D editing through Chain-of-Mesh (CoM), an iterative refinement mechanism grounded in multimodal prompting. In CoM, no additional rendering is required; instead, the original image latent used to generate the 3D mesh is reused and combined with a new editing prompt (e.g., “add crown”, “change color to blue”). This visual-textual pair is then fed into BAGEL’s Qwen module, which generates an updated image latent that encodes the desired modification. The Mesh Head translates this latent into a refined conditioning signal for Hunyuan3D, producing an edited mesh that aligns with the new instruction. This latent–prompting–regeneration loop operates entirely at inference time without any parameter updates, enabling intuitive, language-driven 3D editing through repeated calls to the unified generation-understanding pipeline.

Finally, to support high-level semantic tasks such as captioning or attribute reasoning, UniMesh incorporates a self-reflection module inspired by recent advances in language-agent reasoning. By simulating an Actor–Evaluator–Self-reflection triad, the system can iteratively improve its understanding outputs based on self-diagnosed errors, further strengthening the coupling between perception and generation. 

Together, these components realize a holistic 3D intelligence system: generation informs understanding through realistic 3D hypotheses, while understanding guides generation toward structurally and semantically coherent results. This synergy is the cornerstone of UniMesh’s design and enables capabilities beyond the reach of fragmented, single-task models. 

\subsection{Supervised Fine-Tuning} 

To enable seamless integration between the generative and understanding pathways in UniMesh, we introduce a dedicated Mesh Head module that unifies the FLUX \cite{labs2025flux1kontextflowmatching} decoder from BAGEL and the DINOv2 \cite{oquab2023dinov2} conditioner from Hunyuan3D. This module is designed to directly map the image latent — produced by Qwen within BAGEL’s diffusion pipeline — into a conditioning latent compatible with Hunyuan3D’s shape generation backbone, thereby bypassing the intermediate RGB image reconstruction step and mitigating associated information loss.

We fine-tune the Mesh Head in a supervised manner using the Cap3D dataset. Specifically, for each 3D asset, we select the best view from multi-view images using DiffuRank \cite{luo2024view} and apply photorealistic shadow augmentation to better match the visual characteristics (e.g., cast shadows) commonly present in BAGEL-generated images. These augmented views are then encoded using BAGEL’s FLUX \cite{labs2025flux1kontextflowmatching} image encoder to obtain image latents \( z_{\text{img}} \).

During training, \( z_{\text{img}} \) is fed into the Mesh Head, which outputs a conditioning latent \( z_{\text{cond}} \). This latent is subsequently passed to Hunyuan3D’s shape generation module to produce a signed distance field (SDF) prediction. To supervise this process, we align the ground-truth point cloud \( \mathcal{P}_{\text{gt}} \) with the predicted SDF using the GeDi \cite{Poiesi2021Gedi} alignment algorithm, which establishes a geometric correspondence between surface samples and the implicit field. We then sample points from \( \mathcal{P}_{\text{gt}} \) within the SDF volume and compute a point-to-SDF loss between the sampled ground-truth geometry and the SDF values at those locations. This loss drives end-to-end optimization of the Mesh Head, ensuring that the generated conditioning latent preserves essential 3D structural cues while remaining compatible with the downstream generative pipeline.

\subsection{Chain-of-Mesh}
While traditional 3D generative models produce meshes in a single forward pass, they offer no mechanism for iterative refinement or user-guided editing. To address this, we introduce Chain-of-Mesh (CoM) — a prompting-based iterative framework that leverages UniMesh’s unified architecture to support semantic 3D editing.

\begin{figure}[htbp]
	\centering
	\includegraphics[width=\textwidth]{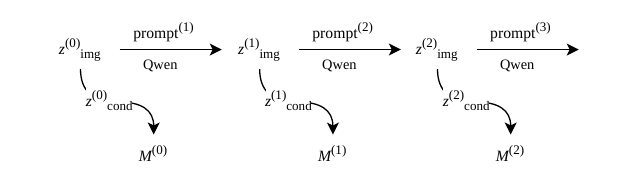}
	\caption{\textbf{Chain of Mesh.} A closed-loop "latent, prompting, and re-generation" cycle.}
	\label{fig:CoM}
\end{figure}

Given an initial image latent \( z_{\text{img}}^{(0)} \) by Qwen, the Mesh Head produces a conditioning latent \( z_{\text{cond}}^{(0)} \), which Hunyuan3D decodes into an initial mesh \( \mathcal{M}^{(0)} \). To edit this mesh, the user provides a new textual instruction (e.g., “make it red,” “add wings”). CoM then feeds the initial image latent \( z_{\text{img}}^{(0)}\) with the new editing prompt into BAGEL’s Qwen module. Qwen jointly interprets the visual context from the reference image latent of the current mesh and the textual instruction, and generates a refined image latent \( z_{\text{img}}^{(1)} \) that encodes the desired modification. This latent is passed through the same Mesh Head to obtain an updated conditioning latent \( z_{\text{cond}}^{(1)} \), which Hunyuan3D uses to produce the edited mesh \( \mathcal{M}^{(1)} \). This process can be repeated, forming the Chain-of-Mesh, as illustrated in \cref{fig:CoM}.

\begin{figure}[htbp]
	\centering
	\includegraphics[width=\textwidth]{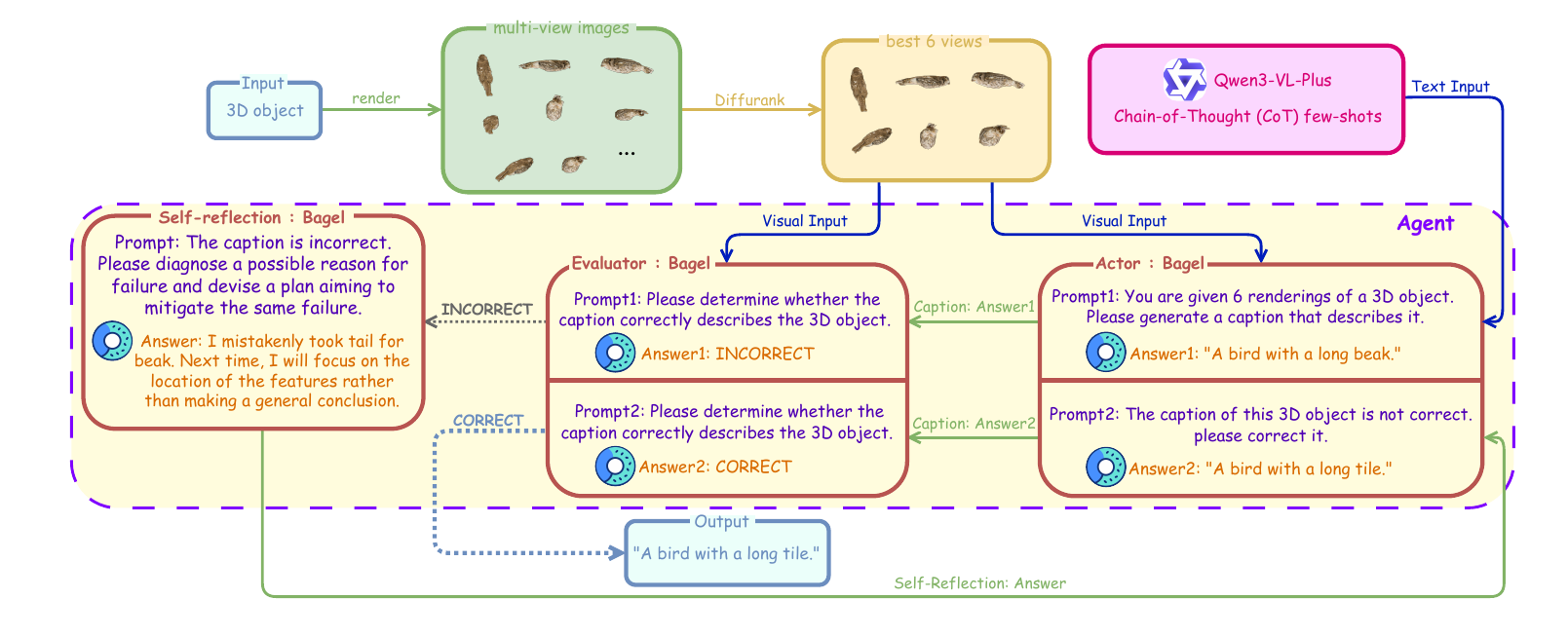}
	\caption{\textbf{Pipeline of Self-Reflection.} The pipeline progresses from a 3D object, through rendering, view selection, to model captioning. The Reflexion agent continuously corrects errors through iterative loops, proposes improvements, and eventually provides the final answer.   }
	\label{fig:reflexion}
\end{figure}

Critically, CoM requires no parameter updates — editing is achieved purely through re-prompting the frozen BAGEL and Hunyuan3D components. By grounding iterative refinement in multimodal prompting rather than explicit error analysis, CoM enables intuitive, language-driven 3D editing while preserving the strengths of the underlying generative and understanding modules.

\subsection{Self-Reflection}

To enhance the model’s capability for comprehensive 3D object understanding, we build UniMesh upon the Reflexion \cite{shinn2023reflexionlanguageagentsverbal} framework, which introduces verbal reinforcement learning through self-reflective feedback. The pipeline of UniMesh for understanding is shown in \cref{fig:reflexion}. 

For each 3D object, we first render multiple views from diverse camera angles to obtain a richer visual representation. We then apply the DiffuRank \cite{luo2024view} method for view selection, enabling the system to automatically identify the six most informative renderings based on view-quality scores. These selected images serve as the visual input for our model.

For the Reflexion components, we employ the Qwen3-VL-Plus model to generate high-quality Chain-of-Thought (CoT) \cite{wei2022chain} few-shot examples that guide subsequent reasoning. Following the standard Reflexion paradigm, instances of the Bagel model are simultaneously used as the Actor, Evaluator, and Self-reflection modules. The Actor produces an initial caption for the 3D object conditioned on the six renderings and CoT demonstrations. The Evaluator then assesses the generated caption and outputs a correctness judgment. If the caption is deemed correct, it is accepted as the final result. If incorrect, we forward the failure-related information — incorrect captions, and corresponding renderings — to the Self-reflection module.

\begin{figure}[htbp]
	\centering
	\includegraphics[width=\textwidth]{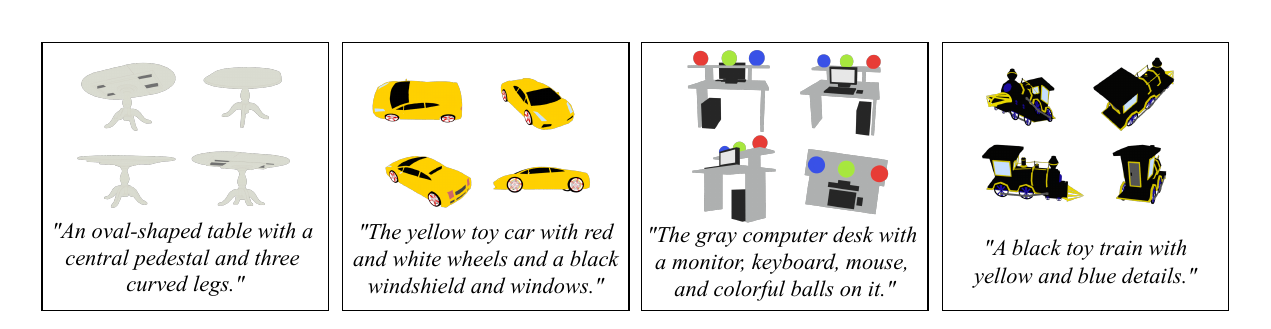}
	\caption{\textbf{Captions generated by UniMesh.} In each box, there are 4 good views of a 3D object and a caption of it generated by UniMesh. UniMesh generates detailed, attribute-rich captions, describing not only object identity but also color combinations, structural elements, etc. }
	\label{fig:captions}
\end{figure}

The Self-reflection module synthesizes these signals into a structured verbal reflection that identifies the specific shortcomings of the Actor’s generation and proposes actionable improvements. This reflection is appended to the Actor’s episodic memory and used as additional context in the next inference cycle. Guided by this self-improvement signal, the Actor regenerates a refined caption. Through this iterative loop of acting, evaluation, and self-reflection, the system progressively improves its captioning ability without parameter updates. Some caption examples are presented in \cref{fig:captions}.

\section{Experiment}

\subsection{Dataset and Metrics}

\subsubsection{3D Object Generation.}

To train and fine-tune the Mesh Head module of UniMesh — which bridges BAGEL’s image latent space with Hunyuan3D’s shape generation backbone — we employ the Cap3D dataset, a large-scale, high-quality 3D asset collection. For each 3D model, we select the most informative viewpoint using DiffuRank \cite{luo2024view}. This ensures that the training signal is both visually representative and geometrically discriminative, facilitating robust generalization to unseen object categories. 

In addition to supervised fine-tuning, we evaluate UniMesh’s capability for text-to-3D generation using prompts derived from the DreamFusion \cite{poole2022dreamfusion}. Specifically, we adopt a curated set of 404 deduplicated text prompts originally used in prior work by Han et al. \cite{han2024flex3d} to ensure fair and reproducible comparisons across methods. These prompts cover diverse object categories, attributes, and compositional descriptions enabling comprehensive assessment of both semantic fidelity and structural plausibility. 

For quantitative evaluation, we report two complementary metrics. (a) CLIP \cite{radford2021learning_clip} Image-Text Similarity: We render multiple views of each generated mesh under consistent lighting and camera settings, then compute the average cosine similarity between the CLIP image embeddings and the text embedding of the input prompt. (b) ViCLIP \cite{wang2023internvid_viclip} Text Similarity: To capture temporal and multi-view consistency, we generate orbital video sequences around each 3D object and measure alignment with the prompt via ViCLIP, which encodes spatiotemporal semantics.

These metrics provide a zero-shot, reference-free assessment of semantic alignment — crucial for evaluating generative models where ground-truth 3D geometry may not be available or meaningful (e.g., for imaginative or composite prompts).

\subsubsection{3D Object Understanding.}
For 3D object captioning task, we use a subset of Cap3D \cite{luo2023scalable} dataset as the test dataset, which includes 3186 3D objects. Each object has a human ground-truth caption. We use several evaluation metrics to quantificationally measure the quality of the captions generated by the models, which include CLIP Image-Text similarity \cite{radford2021learningtransferablevisualmodels}, CLIP Text-Text \cite{shen2021clipbenefitvisionandlanguagetasks}, FID Score \cite{heusel2018ganstrainedtimescaleupdate}, retrieval metrics R@1/5/10 \cite{fang2015captionsvisualconcepts}, and lexical similarity \cite{liu2021swintransformerhierarchicalvision}.

\subsection{Implementation Details}

\subsubsection{3D Object Generation.}

The Mesh Head is initialized by combining the FLUX \cite{labs2025flux1kontextflowmatching} decoder from BAGEL and the DINOv2 \cite{oquab2023dinov2} conditioner from Hunyuan3D-2, where the output of the former serves as the input to the latter. During fine-tuning, we adopt LoRA (Low-Rank Adaptation) \cite{hu2022lora} to efficiently update only the query and value projections within the Mesh Head, with rank \( r = 4 \) and \( \alpha = 8 \), as we empirically found this configuration yields optimal performance while maintaining parameter efficiency. For training, we use the best single view per 3D asset as ranked by DiffuRank \cite{luo2024view}.

To better align real 3D renderings with the visual style of diffusion-generated images, we apply two photorealistic augmentations during training: (a) Drop Shadow Augmentation: We synthesize cast shadows by offsetting the object’s alpha mask, applying Gaussian blur, and compositing a semi-transparent black shadow beneath the object. The shadow offset is randomized in both x and y directions to simulate varying light directions. (b) Shallow Gradient Background: For originally transparent backgrounds, we generate a subtle radial gradient centered near the image middle (with random offsets) using near-white colors. The gradient is smoothed with a large Gaussian blur and combined with low-amplitude Gaussian noise to mimic natural lighting falloff, ensuring the background remains visually unobtrusive while breaking pure transparency. These augmentations collectively mimic the soft shadows and gentle illumination commonly found in diffusion-based image generators, thereby reducing domain gap between Cap3D renderings and BAGEL’s latent space.

Consequently, the Mesh Head not only mitigates information loss from intermediate RGB reconstruction but also learns to disentangle and suppress shadow and background cues in the latent space, focusing instead on core geometric and semantic content.

For supervision, we align the ground-truth point cloud \( \mathcal{P}_{\text{gt}} \) with the predicted SDF as follows: first, we sample a point cloud \( \mathcal{P}_{\text{pred}} \) from the SDF; then, we treat \( \mathcal{P}_{\text{pred}} \) as the reference and apply the GeDi \cite{Poiesi2021Gedi} algorithm to transform \( \mathcal{P}_{\text{gt}} \) into the same scale and orientation. We further refine this alignment using ICP (Iterative Closest Point). Finally, we compute the point-to-SDF loss between the transformed \( \mathcal{P}_{\text{gt}} \) and the predicted SDF.

To reduce memory consumption during training, we employ the Hunyuan3D-2 Mini Turbo variant in the shape generation module and enable FlashVDM \cite{lai2025unleashing}. At inference time, however, we switch back to the full standard Hunyuan3D-2 model to ensure high-fidelity mesh generation.

\subsubsection{3D Object Understanding.}

We evaluate a range of open-source models by generating captions for the 3D objects in our test set and computing standard captioning metrics against human-annotated ground-truth captions. For all models, each 3D object is represented by six rendered views selected through the DiffuRank method \cite{luo2024view}, ensuring consistent and informative visual input. Notably, UniMesh is assessed under its full configuration, incorporating both CoT few-shot and the Reflexion strategy. This setup allows us to comprehensively examine the benefits introduced by our unified understanding framework.

\subsection{Main Results}

\begin{table}[tb]
  \caption{\textbf{Comparison on 3D Object Captioning.} The experiment is conducted on a subset of Cap3D including 3186 3D objects.  
  }
  \label{tbl:obj_cap_results}
  \resizebox{\linewidth}{!}{
		\begin{tabular}{cccccccc}
			\toprule[1.5pt]
			\textbf{Model} & \textbf{CLIP Image-Text$\uparrow$} & \textbf{CLIP Text-Text $\uparrow$} & \textbf{FID$\downarrow$} & \textbf{R@10(\%)$\uparrow$} & \textbf{R@5(\%)$\uparrow$} & \textbf{R@1(\%)$\uparrow$} & \textbf{Lexical Sim.$\uparrow$} \\
			\midrule[1pt]
			Cap3D \cite{luo2023scalable} & 0.287 & 0.694 & 0.123 & \textbf{41.27} & \textbf{33.52} & \textbf{17.51} & 0.158\\
			DiffuRank \cite{luo2024view} & 0.291 & 0.680 & 0.137 & 38.61 & 30.82 & 16.07 & 0.133\\
			LLaVA-3D-7B \cite{zhu2024llava} & 0.278 & 0.698 & 0.209 & 20.34 & 15.57 & 7.19 & 0.146 \\
			Bagel \cite{deng2025bagel} & 0.299 & 0.664 & 0.150 & 35.06 & 28.28 & 14.63 & 0.145 \\
			Qwen2.5-VL-3B \cite{Qwen2.5-VL}  & 0.299 & 0.661 & 0.199 & 34.78 & 27.81 & 13.25 & 0.121\\
			Qwen2.5-VL-7B \cite{Qwen2.5-VL}  & 0.296 & \textbf{0.715} & 0.185 & 32.33 & 25.77 & 12.21 & 0.133\\ 
			Phi-4-multimodel \cite{microsoft2025phi4minitechnicalreportcompact}  & 0.285 & 0.708 & 0.172 & 24.48 & 19.46 & 9.45 & \textbf{0.172}\\
			Kimi-VL \cite{kimiteam2025kimivltechnicalreport} & \textbf{0.312} & 0.597 & 0.213 & 34.93 & 27.90 & 13.87 & 0.116\\
			InternVL3.5-4B \cite{wang2025internvl3_5} & 0.295 & 0.658 & 0.149 & 32.74 & 25.71 & 12.84 & 0.160 \\
			\midrule[1pt]
			\textbf{UniMesh (Ours)}  & 0.297 & 0.686 & \textbf{0.113} & 35.97 & 28.09 & 13.72 & 0.155\\
			\bottomrule[1.5pt]
		\end{tabular}
	}
\end{table}

\subsubsection{Object Captioning.}

The quantitative results are summarized in \cref{tbl:obj_cap_results}. Overall, our UniMesh model demonstrates strong and well-balanced performance across all evaluation metrics. In particular, when considering the weighted aggregation of all metrics, UniMesh demonstrates robust and well-rounded performance, achieving strong CLIP-based semantic alignment (CLIP Image-Text: 0.297, CLIP Text-Text: 0.686) while maintaining a low FID score (0.113), reflecting the high fidelity and naturalness of the generated captions.

Although Cap3D exhibits the highest retrieval accuracy (R@10: 41.27\%), UniMesh achieves a favorable trade-off between retrieval performance (R@10: 35.97\%) and text–image alignment while substantially outperforming most models in visual–textual consistency and caption realism. Moreover, UniMesh consistently maintains competitive lexical similarity, further demonstrating the descriptive quality of its generated captions.

Taken together, these results show that UniMesh offers a robust and well-rounded solution for 3D object captioning, striking an effective balance between semantic relevance, retrieval capability, and caption generation fidelity.

\subsubsection{Text-to-Object Generation.}

\begin{wraptable}{r}{0.5\linewidth}
\vspace{-5pt}
\caption{\textbf{Comparison on Text-to-Object Generation.} UniMesh achieves competitive performance across both CLIP and ViCLIP metrics, outperforming several recent methods including InstantMesh, LGM, and Flex3D.}
\label{tbl:text_to_object}
\centering
\resizebox{\linewidth}{!}{
\begin{tabular}{@{}ccc@{}}
\toprule[1.5pt]
\textbf{Method} & \textbf{CLIP Image-Text $\uparrow$} & \textbf{ViCLIP Text $\uparrow$} \\
\midrule[1pt]
OpenLRM~\cite{openlrm} & 0.243 & 0.229 \\
VFusion3D~\cite{han2024vfusion3d} & 0.265 & 0.238 \\
LGM~\cite{tang2024lgm} & 0.266 & 0.240 \\
InstantMesh~\cite{xu2024instantmeshefficient3dmesh} & 0.272 & 0.236 \\
GRM~\cite{xu2024grm} & 0.268 & 0.253 \\
LN3Diff~\cite{lan2024ln3diff} & 0.252 & 0.234 \\
3DTopia-XL~\cite{chen2025primx} & 0.254 & 0.231 \\
Flex3D~\cite{han2024flex3d} & 0.277 & \textbf{0.255} \\
\midrule[1pt]
\textbf{UniMesh (Ours)} & \textbf{0.296} & 0.243 \\
\bottomrule[1.5pt]
\end{tabular}
}
\vspace{-10pt}
\end{wraptable}

We evaluate UniMesh's capability of text-to-3D generation using prompts from DreamFusion \cite{poole2022dreamfusion}, following the evaluation protocol of Han et al. \cite{han2024flex3d} As shown in \cref{tbl:text_to_object}, UniMesh achieves competitive performance across both CLIP and ViCLIP metrics, outperforming several recent methods including InstantMesh \cite{xu2024instantmeshefficient3dmesh}, LGM \cite{tang2024lgm}, and Flex3D \cite{han2024flex3d}. 

Notably, UniMesh attains a CLIP Image-Text similarity score of 0.296, surpassing all compared baselines and setting a new state-of-the-art among open models evaluated under this protocol. This strong semantic alignment reflects UniMesh’s ability to generate 3D meshes that closely match the textual description — not just in coarse shape, but also in fine-grained attributes such as color, pose, and compositional details (e.g., “an astronaut in blue spacesuit holding the Moon.”), as shown as \cref{fig:teaser} . 

This high level of semantic fidelity is largely attributed to the Qwen backbone within BAGEL, which provides rich, multimodal understanding of natural language prompts. By directly mapping Qwen’s image latent into Hunyuan3D’s shape conditioning space via the Mesh Head, UniMesh inherits Qwen’s powerful linguistic grounding while bypassing the information loss inherent in intermediate RGB rendering. As a result, the generated 3D content preserves the semantic intent of the prompt more faithfully than methods relying solely on visual diffusion priors or category-specific reconstruction. 

While other methods (e.g., OpenLRM, 3DTopia-XL) achieve comparable scores, they often require larger models or specialized training data. In contrast, UniMesh leverages existing large-scale vision-language capabilities in a unified, efficient architecture — demonstrating that semantic-aware 3D generation can be achieved through intelligent integration rather than brute-force scaling.

\begin{table}[tb]
  \caption{\textbf{3D Object Captioning Ablation results.} The ablation study is conducted on a subset of Cap3D including 200 3D objects. The full configuration (DiffuRank + CoT + Reflexion) achieves the strongest results across most metrics. 
  }
  \label{tbl:obj_cap_ablation}
  \resizebox{\linewidth}{!}{
		\begin{tabular}{cccccccc}
			\toprule[1.5pt]
			\textbf{Method} & \textbf{CLIP Image-Text $\uparrow$} & \textbf{CLIP Text-Text $\uparrow$} & \textbf{FID$\downarrow$} & \textbf{R@10(\%)$\uparrow$} & \textbf{R@5(\%)$\uparrow$} & \textbf{R@1(\%)$\uparrow$} & \textbf{Lexical Sim.$\uparrow$} \\
			\midrule[1pt]
			No Diffurank & 0.299 & 0.657 & 0.382 & 69.00 & 58.00 & 35.50 & 0.153\\
			DiffuRank + no CoT & \textbf{0.302} & 0.669 & 0.385 & 67.50 & 56.50 & \textbf{37.00} & 0.141\\  
			DiffuRank + CoT & 0.298 & 0.693 & \textbf{0.345} & \textbf{69.50} & \textbf{63.00} & 35.50 & 0.159\\       
			\midrule[1pt]
			\textbf{Ours}  & 0.298 & \textbf{0.694} & \textbf{0.345} & \textbf{69.50} & \textbf{63.00} & 35.50 & \textbf{0.160} \\
			\bottomrule[1.5pt]
		\end{tabular}
	}
\end{table}

\subsubsection{Object Editing.}

Beyond static generation, UniMesh demonstrates strong capabilities in semantic-aware 3D object editing, enabling users to iteratively refine or modify generated meshes based on natural language instructions — all without requiring manual mesh manipulation or retraining. 

As illustrated in \cref{fig:teaser}, given an initial prompt (top row), UniMesh first generates a high-fidelity 3D mesh. The user can then provide a simple textual edit instruction (bottom row), such as: (a) changing color: “blue motorcycle” → “red motorcycle” , (b) adding attributes: “astronaut” → “astronaut holding the Moon” or “wooden lion” → “lion with crown” , (c) modifying structure: “bulldozer with tracks” → “bulldozer with wheels” and (d) removing object: "flowers" → "one flower".

The resulting edited meshes preserve the original geometry while accurately incorporating the requested semantic changes — visually indistinguishable from objects designed from scratch for that description. 

This capability stems from UniMesh’s unified architecture: by feeding reference image latent of the current mesh alongside the new prompt back into BAGEL’s Qwen module, the system effectively “understands” what needs to be changed and generates a refined conditioning latent that guides Hunyuan3D toward the desired edit. Crucially, this process requires no fine-tuning, no explicit mesh deformation, and no additional training data — making it highly accessible for interactive design or rapid prototyping. 

Compared to prior text-guided editing methods that rely on optimization loops or conditional diffusion models, UniMesh achieves more intuitive, immediate, and semantically coherent edits — directly leveraging the linguistic grounding of Qwen to interpret instructions at the object level. The results in \cref{fig:teaser} showcase not just technical feasibility, but also the practical value of such a system for creative workflows where users wish to iterate on 3D content using natural language.

\subsection{Ablation Study}

To better quantify the contribution of each component in our Self-Reflection framework, we conduct an ablation study on a subset of 200 3D objects, using a model without DiffuRank, without CoT, and without Reflexion as the baseline. We progressively introduce three key modules—view selection, Chain-of-Thought (CoT) guidance, and the Reflexion agent—to observe their individual and cumulative effects on 3D object understanding.

First, incorporating DiffuRank for view selection leads to noticeable improvements in CLIP alignment and retrieval accuracy, indicating that selecting informative views enhances the model’s grounding in 3D geometry. Next, adding CoT few-shot examples further boosts semantic alignment—especially CLIP Text-Text—and significantly reduces FID, demonstrating that structured reasoning helps the model generate more coherent and descriptive captions. Finally, introducing the Reflexion module produces an additional gain, particularly in lexical similarity, suggesting that iterative self-refinement enables the model to capture more nuanced attributes of 3D objects.

As summarized in \cref{tbl:obj_cap_ablation}, each component contributes positively to overall performance, and the full configuration (DiffuRank + CoT + Reflexion) achieves the strongest results across most metrics. These findings validate the effectiveness of our design and highlight the complementary roles of view selection, CoT few-shot, and self-reflective learning.

\section{Limitation and Future Work}

Although Chain-of-Mesh enables semantic-aware editing, it relies on reference image latent rather than direct 3D understanding. Similarly, the Reflexion framework uses a BAGEL-based evaluator, whose limited 3D reasoning capability can lead to incorrect judgments and degrade self-reflection quality. 

Future work includes:
(1) training models to understand 3D objects directly in their native geometric representation;
(2) improving the Reflexion framework with more reliable evaluation and reflection mechanisms.

\section{Conclusion}

We have presented \textbf{UniMesh}, a unified 3D vision framework that bridges the long-standing divide between 3D generation and understanding. By integrating BAGEL and Hunyuan3D through a purpose-built Mesh Head, UniMesh enables direct latent-space transfer from semantic-rich image representations to high-fidelity 3D shape generation—bypassing intermediate rendering and preserving geometric integrity. 

Beyond one-pass synthesis, UniMesh introduces two key reasoning mechanisms:
(i) a self-reflection module that iteratively critiques and refines understanding outputs (e.g., captions) through verbal feedback, and
(ii) Chain-of-Mesh (CoM), which enables zero-shot, text-guided 3D editing by transforming semantic critiques into actionable geometric refinements. 

Together, these components realize a closed-loop system where generation informs understanding and understanding guides generation—yielding mutual improvement without task-specific redesign. While current limitations include reliance on 2D views for 3D reasoning and evaluator imperfections in reflection, UniMesh establishes a foundation for holistic 3D intelligence. We hope this work inspires future efforts toward truly integrated, self-improving 3D vision systems.

\clearpage  %

\bibliographystyle{splncs04}
\bibliography{main}

\clearpage

\appendix

\section{DETAILED CAPTIONS QUALITY ANALYSIS}

As shown in \cref{tbl:obj_cap_results}, the evaluation results for 3D object captioning reveal distinct performance characteristics among various models. Among the 3D-specialized models, Cap3D achieves the best overall performance, particularly in retrieval tasks (R@10: 41.27\%, R@5: 33.52\%, R@1: 17.51\%), demonstrating the effectiveness of large-scale 3D-text pair pretraining for understanding and describing 3D objects. DiffuRank follows closely with competitive CLIP scores (Image-Text: 0.291, Text-Text: 0.680) and retrieval metrics, benefiting from its diffusion-based architecture that selects best views, though slightly trailing Cap3D in lexical similarity (0.133 vs. 0.158). LLaVA-3D-7B, while leveraging a strong multimodal foundation, shows relatively weaker retrieval performance (R@10: 20.34\%) and higher FID (0.209), indicating limitations in aligning 3D visual features with textual descriptions despite its innovative fusion design.

General-purpose vision-language models (VLMs) exhibit diverse strengths. Kimi-VL stands out in CLIP Image-Text similarity (0.312), suggesting superior cross-modal alignment, though its lower Text-Text similarity (0.597) and lexical similarity (0.116) imply challenges in maintaining semantic consistency. Qwen2.5-VL models demonstrate balanced performance: the 7B variant achieves the highest CLIP Text-Text score (0.715), reflecting strong semantic understanding, while the 3B variant shows competitive retrieval metrics (R@10: 34.78\%) and lexical similarity (0.121), highlighting efficiency in resource-constrained settings. Phi-4-multimodel excels in lexical similarity (0.172) and maintains solid CLIP scores, indicating its proficiency in domain-specific terminology. InternVL3.5-4B achieves a low FID (0.149) and good lexical similarity (0.160), benefiting from its advanced vision-language interaction mechanisms. Bagel, despite being less prominent, performs reasonably well across metrics.

Notably, open-source models demonstrate competitive capabilities, often rivaling or surpassing specialized models in certain dimensions. For instance, Qwen2.5-VL-7B’s Text-Text CLIP score (0.715) exceeds all other models, showcasing the strength of large-scale pretraining. Phi-4-multimodel’s lexical similarity (0.172) outperforms even Cap3D, indicating its potential for tasks requiring precise terminology usage.

Our proposed UniMesh achieves the best FID score (0.113), significantly outperforming all other models, which underscores its superior capability in generating high-quality captions that closely match ground-truth distributions. It also attains competitive performance in retrieval tasks (R@10: 35.97\%, R@5: 28.09\%, R@1: 13.72\%) and CLIP metrics (Image-Text: 0.297, Text-Text: 0.686), demonstrating a balanced trade-off between generation fidelity and semantic alignment. This success stems from our self-reflection mechanism, which is built on Bagel model.

Comparing models of different scales, larger models like Qwen2.5-VL-7B and Kimi-VL tend to excel in CLIP-based metrics, but smaller models such as Qwen2.5-VL-3B and InternVL3.5-4B still deliver competitive retrieval results and lower FID, validating their practicality in scenarios with limited computational resources. Overall, the results highlight that task-specific design and hybrid training strategies, as embodied by UniMesh, can achieve excellent generation quality while maintaining robust retrieval and semantic understanding capabilities.

\end{document}